\pdfoutput=1

\documentclass[11pt]{article}

\usepackage{acl}

\usepackage{times}
\usepackage{latexsym}
\usepackage{enumitem}

\usepackage[T1]{fontenc}

\usepackage[utf8]{inputenc}

\usepackage{microtype}

\title{Automatic Exploration of Textual Environments with Language-Conditioned Autotelic Agents}

\author{
  Laetitia Teodorescu \\
  Inria \\
  \texttt{laetitia.teodorescu@inria.fr} \\\And
  Xingdi Yuan \\
  Microsoft Research \\
  \texttt{eric.yuan@microsoft.com} \\\AND
  Marc-Alexandre Côté \\
  Microsoft Research \\
  \texttt{macote@microsoft.com} \\\And
  Pierre-Yves Oudeyer \\
  Inria \\
  \texttt{pierre-yves.oudeyer@inria.fr}
  }

\begin{document}
\maketitle

\newcounter{para}
\newcommand\mypara{\par\refstepcounter{para}\textbf{\thepara.}\space}




This extended abstract discusses the opportunities and challenges of studying intrinsically-motivated agents for exploration in textual environments.

Humans begin their life with very few skills, and over the course of only a few years learn complex motor coordination and locomotion capabilities, begin mastering vocalization and language, and form a rich model of their physical and social surroundings. One of the main drivers of this phenomenal knowledge acquisition is intrinsically-motivated exploration \cite{Oudeyer2007}, for instance through exploratory play \cite{schulz-play-2020, davidson_gureckis_lake_2022}. The developmental perspective on AI tries to emulate this exploratory behavior in artificial agents to achieve mastery of diverse and complex repertoires of skills \cite{forestier2017intrinsically}. When placed in open-ended environments, a successful intrinsically motivated agent will explore the space of interesting and diverse outcomes, ignoring random and unachievable subspaces of the world, reusing its previously acquired skills as stepping stones \cite{greatness} to discover new ones.

One possible implementation of exploration in RL agents are so-called autotelic agents \cite{colas:tel-03337625}, that is, goal-conditioned Reinforcement Learning (RL) agents operating in rewardless environments that are able to choose what goal to pursue. In this case, the reward is given by a goal-satisfaction function and not extrinsically by the environment. Goal-conditioned policies have been extensively studied in the case of extrinsic goals \cite{schaul2015universal}. In the case of intrinsically chosen goals, the goal-selection mechanism allows autotelic agents to form a self-curriculum, progressing from easier to increasingly harder goals until all achievable skills have been mastered. In this perspective, the goal representation is of paramount importance. Most previous works (for instance \citet{andrychowicz2017hindsight}) have used concrete end-state representations such as raw observations, images or embeddings, which has some drawbacks. A goal should be insensitive to changes in the environment that are uncontrollable (such as the color of the sky), to avoid the agent targeting impossible goals (for instance changing the sky color), or to provide useful abstraction for goal achievement (such as considering the goal of navigating to the garden is satisfied regardless of sky color). Furthermore, the agent should ideally be able to combine known goals into novel ones. \textbf{Goals expressed as language} \cite{tam2022semantic, colas2020language, mu2022improving} fulfill both conditions: they are at once abstract and combinatorial \cite{sep-compositionality}; they are thus a prime way for autotelic agents to self-specify goals to be executed in the environment.

\section{A bridge between autotelic agents and text environments}

The main point of this essay is the relevance of studying language autotelic agents in textual environments \cite{cote2018textworld, hausknecht2020interactive, wang2022scienceworld}, both for testing exploration methods in a context that is at once simple experimentally and rich from the perspective of environment interactions; and for transferring the skills of general-purpose agents trained to explore in an autonomous way to the predefined tasks of textual environment benchmarks. We identify three key properties, plus one additional benefit, of text worlds:

\vspace{5pt}

\mypara \textbf{Depth of learnable skills:} skills learnable in the world should involve multiple low-level actions and be nested, such that mastering one skill opens up the possibility of mastering more complex skills. Interactive fiction (IF) \cite{hausknecht2020interactive} games usually feature an entire narrative and extensive maps, such that navigating and passing obstacles requires many successful actions (and subgoals) to be completed. While the original TextWorld levels were not as deep as would be desirable, other non-IF text worlds such as ScienceWorld feature nested repertoires of skills (such as learning to navigate to learn to grow plants to learn the rules of Mendelian genomics);

\mypara \textbf{Breadth of the world:} there should be many paths to explore in the environment; this ensures that we train agents that are able to follow a wide diversity of possible goals, instead of learning to achieve goals along a linear path. This allows us to study generally-capable agents. Some IF games are very linear, having a clear progression from start to finish (e.g., \href{https://ifdb.org/viewgame?id=tqvambr6vowym20v}{Acorn Court}, \href{https://ifdb.org/viewgame?id=1po9rgq2xssupefw}{Detective}; others have huge maps that an agent has to explore before it can progress in the quest (e.g., \href{https://ifdb.org/viewgame?id=0dbnusxunq7fw5ro}{Zork}, \href{https://ifdb.org/viewgame?id=ouv80gvsl32xlion}{Hitchhiker's Guide to the Galaxy}). Exploration heuristics are a part of some successful methods for playing IF with RL \cite{yao2020keep}. ScienceWorld \cite{wang2022scienceworld} has an underlying physical engine allowing for a combinatorial explosion of possibilities like making new objects, combining existing objects, changing states of matter, etc.

\mypara \textbf{Niches of progress}: real-world environments have both easy skills and unlearnable skills. Our simulated environments should mimic this property to test the agent's ability to focus only on highly learnable parts of the space and avoid spending effort on uncontrollable aspects of the environment. In textual environments, high depth implies that some skills are much more learnable than others, already implementing some progress niches. The combinatorial property of language goals allows us to define many unfeasible goals, goals that an autotelic agent has to avoid spending too many resources on.

\mypara \textbf{Language representation for goals:} a language-conditioned agent has to learn to ground its goal representation in its environment \cite{harnad1990symbol, hill2020grounded}, to know when a given observation or sequence of observations satisfies a given goal, or to know what goals were achieved in a given trajectory. This grounding is made much simpler in environments with a single modality; relating language goals to language observations is simpler than grounding language in pixels or image embeddings. This allows us to study language-based exploration in a simpler context.



\vspace{5pt}
\setcounter{para}{0}

\section{Drivers of exploration in autotelic agents}

We identify three main drivers of exploration in autotelic agents. Environments we use should support exploration algorithms that implement these principles; the resulting agents then have a chance to acquire a diverse set of skills that can be repurposed for solving the benchmarks proposed by textual environments.


\vspace{5pt}


\mypara \textbf{Goal self-curriculum:} automatic goal selection \cite{portelas2020automatic} allows the agent to refine its skills on the edge of what it currently masters. Among metrics used to select goals are novelty/surprise of a goal \cite{tam2022semantic, burda2018exploration}, intermediate competence on goals \cite{campero2020learning}, ensemble disagreement \cite{pathak2019self}, or (absolute) learning progress \cite{colas2019curious}. Progress niches in textual environments support such goal curriculum;
\mypara \textbf{Additional exploration after goal achievement:} after achieving a given goal, the agent continue to run for a time to push the boundary of explored space \cite{ecoffet2021first}. The depth of text worlds makes goal chaining relevant, such that an agent that has achieved a known goal can imagine additional goals to pursue. Random exploration can also be used once a known goal has been achieved. Agents exploring in textual environments and choosing uniformly among the set of valid actions in a given state have a high chance of effecting meaningful changes in the environment, making discovery of new skills probable. This property is relevant in any environment with high depth, and both IF and ScienceWorld fit this description.
\mypara \textbf{Goal composition:} as mentioned above, this means using the compositionality afforded by language goals to imagine novel goals in the environment \cite{colas2020language}. Goal-chaining is an example of composition, but language offers many other composition possibilities, such as recombining known verbs, nouns and attributes in novel ways, or making analogies. This is relevant if there exists some transfer between the skills required to accomplish similar goal constructions (e.g., picking up an apple and picking up a carrot requires very similar actions if both are in the kitchen). This is at least partially true in textual environments where objects of the same type usually have similar affordances.

\vspace{5pt}

\section{Challenges for autotelic textual agents}
\setcounter{para}{0}

Text worlds bring a set of unique challenges for autotelic agents, among which we foresee:

\mypara The goal space can be very large. An agent with a limited training budget needs to focus on a subset of the goal space, possibly discovering only a fraction of what is discoverable within the environment. This calls for finer goal-sampling approaches that encourage the agent at making the most out of its allocated time to explore the environment. In addition, we need better methods to push the agent's exploration towards certain parts of the space (e.g., warm-starting the replay buffer with existing trajectories, providing linguistic common-sense knowledge);

\mypara The action space is also very large in textual environments, making exploration (especially methods based on random action selection) potentially challenging.

\mypara Agents must be trajectory-efficient for a given goal; complex goals might be seen only once;

\mypara Catastrophic forgetting needs to be alleviated, so that learning to achieve new goals does not impair the skills learned previously;

\mypara Partial observability means that agent architectures need to include some form of memory.


\vspace{5pt}

Agents trained in such environments will learn a form of language use, not by predicting the most likely sequence of words from a large-scale dataset \cite{Radford2018ImprovingLU, brown2020language} but by learning to use it pragmatically to effect changes in the environment. Of course, the limits of the autotelic agent's world will mean the limits of its language; an interesting development is to build agents that explore textual environments to refine external linguistic knowledge provided by a pretrained language model. This external knowledge repository can be seen as culturally-accumulated common sense, a perspective that is related to so-called Vygotskian AI \cite{colas:tel-03337625} in which a developmental agent learns by interacting with an external social partner that imparts outside language knowledge and organizes the world so as to facilitate the autotelic agent's exploration.

\vspace{5pt}



To conclude, textual environments are ideal testbeds for autotelic language-conditioned agents, and conversely such agents can bring progress on text world benchmarks. There is also promise in the interaction between exploratory agents and large language models encoding exterior linguistic knowledge. Preliminary steps have been taken in this direction \cite{madotto2020exploration} but the full breadth of drivers of exploration we identify has yet to be studied. We hope to foster discussion, define concrete implementations and identify challenges by bringing together the developmental perspective on AI and the textual environment community.

\bibliography{acl_latex}

\begin{thebibliography}{27}
\expandafter\ifx\csname natexlab\endcsname\relax\def\natexlab#1{#1}\fi

\bibitem[{Andrychowicz et~al.(2017)Andrychowicz, Wolski, Ray, Schneider, Fong,
  Welinder, McGrew, Tobin, Pieter~Abbeel, and
  Zaremba}]{andrychowicz2017hindsight}
Marcin Andrychowicz, Filip Wolski, Alex Ray, Jonas Schneider, Rachel Fong,
  Peter Welinder, Bob McGrew, Josh Tobin, OpenAI Pieter~Abbeel, and Wojciech
  Zaremba. 2017.
\newblock Hindsight experience replay.
\newblock \emph{Advances in neural information processing systems}, 30.

\bibitem[{Brown et~al.(2020)Brown, Mann, Ryder, Subbiah, Kaplan, Dhariwal,
  Neelakantan, Shyam, Sastry, Askell et~al.}]{brown2020language}
Tom Brown, Benjamin Mann, Nick Ryder, Melanie Subbiah, Jared~D Kaplan, Prafulla
  Dhariwal, Arvind Neelakantan, Pranav Shyam, Girish Sastry, Amanda Askell,
  et~al. 2020.
\newblock Language models are few-shot learners.
\newblock \emph{Advances in neural information processing systems},
  33:1877--1901.

\bibitem[{Burda et~al.(2018)Burda, Edwards, Storkey, and
  Klimov}]{burda2018exploration}
Yuri Burda, Harrison Edwards, Amos Storkey, and Oleg Klimov. 2018.
\newblock Exploration by random network distillation.
\newblock \emph{arXiv preprint arXiv:1810.12894}.

\bibitem[{Campero et~al.(2020)Campero, Raileanu, K{\"u}ttler, Tenenbaum,
  Rockt{\"a}schel, and Grefenstette}]{campero2020learning}
Andres Campero, Roberta Raileanu, Heinrich K{\"u}ttler, Joshua~B Tenenbaum, Tim
  Rockt{\"a}schel, and Edward Grefenstette. 2020.
\newblock Learning with amigo: Adversarially motivated intrinsic goals.
\newblock \emph{arXiv preprint arXiv:2006.12122}.

\bibitem[{Chu and Schulz(2020)}]{schulz-play-2020}
Junyi Chu and Laura Schulz. 2020.
\newblock \href {https://doi.org/10.1146/annurev-devpsych-070120-014806} {Play,
  curiosity, and cognition}.
\newblock \emph{Annual Review of Developmental Psychology}, 2.

\bibitem[{Colas(2021)}]{colas:tel-03337625}
C{\'e}dric Colas. 2021.
\newblock \href {https://tel.archives-ouvertes.fr/tel-03337625} {\emph{{Towards
  Vygotskian Autotelic Agents : Learning Skills with Goals, Language and
  Intrinsically Motivated Deep Reinforcement Learning}}}.
\newblock Theses, {Universit{\'e} de Bordeaux}.

\bibitem[{Colas et~al.(2019)Colas, Fournier, Chetouani, Sigaud, and
  Oudeyer}]{colas2019curious}
C{\'e}dric Colas, Pierre Fournier, Mohamed Chetouani, Olivier Sigaud, and
  Pierre-Yves Oudeyer. 2019.
\newblock Curious: intrinsically motivated modular multi-goal reinforcement
  learning.
\newblock In \emph{International conference on machine learning}, pages
  1331--1340. PMLR.

\bibitem[{Colas et~al.(2020)Colas, Karch, Lair, Dussoux, Moulin-Frier, Dominey,
  and Oudeyer}]{colas2020language}
C{\'e}dric Colas, Tristan Karch, Nicolas Lair, Jean-Michel Dussoux, Cl{\'e}ment
  Moulin-Frier, Peter Dominey, and Pierre-Yves Oudeyer. 2020.
\newblock Language as a cognitive tool to imagine goals in curiosity driven
  exploration.
\newblock \emph{Advances in Neural Information Processing Systems},
  33:3761--3774.

\bibitem[{C{\^o}t{\'e} et~al.(2018)C{\^o}t{\'e}, K{\'a}d{\'a}r, Yuan, Kybartas,
  Barnes, Fine, Moore, Hausknecht, Asri, Adada et~al.}]{cote2018textworld}
Marc-Alexandre C{\^o}t{\'e}, Akos K{\'a}d{\'a}r, Xingdi Yuan, Ben Kybartas,
  Tavian Barnes, Emery Fine, James Moore, Matthew Hausknecht, Layla~El Asri,
  Mahmoud Adada, et~al. 2018.
\newblock Textworld: A learning environment for text-based games.
\newblock In \emph{Workshop on Computer Games}, pages 41--75. Springer.

\bibitem[{Davidson et~al.(2022)Davidson, Gureckis, and
  Lake}]{davidson_gureckis_lake_2022}
Guy Davidson, Todd~M Gureckis, and Brenden~M Lake. 2022.
\newblock \href {https://doi.org/10.31234/osf.io/byzs5} {Creativity,
  compositionality, and common sense in human goal generation}.

\bibitem[{Ecoffet et~al.(2021)Ecoffet, Huizinga, Lehman, Stanley, and
  Clune}]{ecoffet2021first}
Adrien Ecoffet, Joost Huizinga, Joel Lehman, Kenneth~O Stanley, and Jeff Clune.
  2021.
\newblock First return, then explore.
\newblock \emph{Nature}, 590(7847):580--586.

\bibitem[{Forestier et~al.(2017)Forestier, Portelas, Mollard, and
  Oudeyer}]{forestier2017intrinsically}
S{\'e}bastien Forestier, R{\'e}my Portelas, Yoan Mollard, and Pierre-Yves
  Oudeyer. 2017.
\newblock Intrinsically motivated goal exploration processes with automatic
  curriculum learning.
\newblock \emph{arXiv preprint arXiv:1708.02190}.

\bibitem[{Harnad(1990)}]{harnad1990symbol}
Stevan Harnad. 1990.
\newblock The symbol grounding problem.
\newblock \emph{Physica D: Nonlinear Phenomena}, 42(1-3):335--346.

\bibitem[{Hausknecht et~al.(2020)Hausknecht, Ammanabrolu, C{\^o}t{\'e}, and
  Yuan}]{hausknecht2020interactive}
Matthew Hausknecht, Prithviraj Ammanabrolu, Marc-Alexandre C{\^o}t{\'e}, and
  Xingdi Yuan. 2020.
\newblock Interactive fiction games: A colossal adventure.
\newblock In \emph{Proceedings of the AAAI Conference on Artificial
  Intelligence}, volume~34, pages 7903--7910.

\bibitem[{Hill et~al.(2020)Hill, Tieleman, von Glehn, Wong, Merzic, and
  Clark}]{hill2020grounded}
Felix Hill, Olivier Tieleman, Tamara von Glehn, Nathaniel Wong, Hamza Merzic,
  and Stephen Clark. 2020.
\newblock Grounded language learning fast and slow.
\newblock \emph{arXiv preprint arXiv:2009.01719}.

\bibitem[{Madotto et~al.(2020)Madotto, Namazifar, Huizinga, Molino, Ecoffet,
  Zheng, Papangelis, Yu, Khatri, and Tur}]{madotto2020exploration}
Andrea Madotto, Mahdi Namazifar, Joost Huizinga, Piero Molino, Adrien Ecoffet,
  Huaixiu Zheng, Alexandros Papangelis, Dian Yu, Chandra Khatri, and Gokhan
  Tur. 2020.
\newblock Exploration based language learning for text-based games.
\newblock \emph{arXiv preprint arXiv:2001.08868}.

\bibitem[{Mu et~al.(2022)Mu, Zhong, Raileanu, Jiang, Goodman, Rockt{\"a}schel,
  and Grefenstette}]{mu2022improving}
Jesse Mu, Victor Zhong, Roberta Raileanu, Minqi Jiang, Noah Goodman, Tim
  Rockt{\"a}schel, and Edward Grefenstette. 2022.
\newblock Improving intrinsic exploration with language abstractions.
\newblock \emph{arXiv preprint arXiv:2202.08938}.

\bibitem[{Oudeyer and Kaplan(2007)}]{Oudeyer2007}
Pierre-Yves Oudeyer and Frederic Kaplan. 2007.
\newblock \href {https://doi.org/10.3389/neuro.12.006.2007} {What is intrinsic
  motivation? a typology of computational approaches}.
\newblock \emph{Frontiers in neurorobotics}, 1:6--6.
\newblock 18958277[pmid].

\bibitem[{Pathak et~al.(2019)Pathak, Gandhi, and Gupta}]{pathak2019self}
Deepak Pathak, Dhiraj Gandhi, and Abhinav Gupta. 2019.
\newblock Self-supervised exploration via disagreement.
\newblock In \emph{International conference on machine learning}, pages
  5062--5071. PMLR.

\bibitem[{Portelas et~al.(2020)Portelas, Colas, Weng, Hofmann, and
  Oudeyer}]{portelas2020automatic}
R{\'e}my Portelas, C{\'e}dric Colas, Lilian Weng, Katja Hofmann, and
  Pierre-Yves Oudeyer. 2020.
\newblock Automatic curriculum learning for deep rl: A short survey.
\newblock \emph{arXiv preprint arXiv:2003.04664}.

\bibitem[{Radford and Narasimhan(2018)}]{Radford2018ImprovingLU}
Alec Radford and Karthik Narasimhan. 2018.
\newblock Improving language understanding by generative pre-training.

\bibitem[{Schaul et~al.(2015)Schaul, Horgan, Gregor, and
  Silver}]{schaul2015universal}
Tom Schaul, Dan Horgan, Karol Gregor, and David Silver. 2015.
\newblock Universal value function approximators.
\newblock page 1312–1320.

\bibitem[{Stanley and Lehman(2015)}]{greatness}
Kenneth~O. Stanley and Joel Lehman. 2015.
\newblock \emph{Why Greatness Cannot Be Planned: The Myth of the Objective}.
\newblock Springer Publishing Company, Incorporated.

\bibitem[{Szabó(2020)}]{sep-compositionality}
Zoltán~Gendler Szabó. 2020.
\newblock {Compositionality}.
\newblock In Edward~N. Zalta, editor, \emph{The {Stanford} Encyclopedia of
  Philosophy}, {F}all 2020 edition. Metaphysics Research Lab, Stanford
  University.

\bibitem[{Tam et~al.(2022)Tam, Rabinowitz, Lampinen, Roy, Chan, Strouse, Wang,
  Banino, and Hill}]{tam2022semantic}
Allison~C Tam, Neil~C Rabinowitz, Andrew~K Lampinen, Nicholas~A Roy,
  Stephanie~CY Chan, DJ~Strouse, Jane~X Wang, Andrea Banino, and Felix Hill.
  2022.
\newblock Semantic exploration from language abstractions and pretrained
  representations.
\newblock \emph{arXiv preprint arXiv:2204.05080}.

\bibitem[{Wang et~al.(2022)Wang, Jansen, C{\^o}t{\'e}, and
  Ammanabrolu}]{wang2022scienceworld}
Ruoyao Wang, Peter Jansen, Marc-Alexandre C{\^o}t{\'e}, and Prithviraj
  Ammanabrolu. 2022.
\newblock Scienceworld: Is your agent smarter than a 5th grader?
\newblock \emph{arXiv preprint arXiv:2203.07540}.

\bibitem[{Yao et~al.(2020)Yao, Rao, Hausknecht, and Narasimhan}]{yao2020keep}
Shunyu Yao, Rohan Rao, Matthew Hausknecht, and Karthik Narasimhan. 2020.
\newblock Keep calm and explore: Language models for action generation in
  text-based games.
\newblock \emph{arXiv preprint arXiv:2010.02903}.

\end{thebibliography}
\bibliographystyle{acl_natbib}

\appendix



\end{document}